\documentclass[runningheads]{llncs}

\usepackage[T1]{fontenc}
\usepackage{graphicx,verbatim}
\usepackage{hyperref}
\usepackage{xcolor}
\usepackage{multirow}
\usepackage{booktabs}
\usepackage{amsfonts}
\usepackage{soul}
\usepackage{xspace}
\usepackage{fontawesome5}
\usepackage{paralist}
\usepackage{ragged2e}
\usepackage[most]{tcolorbox}
\usepackage{wrapfig}

\newtcbox{\senthl}[1]{%
  on line,
  enhanced,
  boxrule=0pt,
  arc=1pt,
  outer arc=1pt,
  left=1pt,right=1pt,top=1pt,bottom=1pt,
  boxsep=0pt,
  colback=#1,
  colframe=#1,
}

\definecolor{myRoyalBlue}{HTML}{4169E1}
\definecolor{myMagenta}{HTML}{FF00FF}
\definecolor{myAquaMarine}{HTML}{7FFFD4}
\definecolor{myPurple}{HTML}{A020F0}
\definecolor{myCoral}{HTML}{FF7F50}
\definecolor{SeaGreen}{HTML}{2E8B57}

\definecolor{assymetry}{cmyk}{0.45, 0., 0.60, 0.}
\definecolor{border}{cmyk}{0.46, 0.1, 0., 0.}
\definecolor{colour}{cmyk}{0.14, 0.21, 0.31, 0.02}
\usepackage{enumitem}

\soulregister\cite7
\soulregister\ref7

\DeclareUnicodeCharacter{2227}{\wedge}

\newcommand{\sys}{\textsc{DermaFlux}\xspace}

\newcommand{\flux}{\textsc{Flux.1}\xspace}

\begin{document}

\title{\sys: Synthetic Skin Lesion Generation with Rectified Flows for Enhanced Image Classification}

\titlerunning{\sys}


\author{Stathis Galanakis\inst{1} \and
Alexandros Koliousis\inst{2} \and
Stefanos Zafeiriou\inst{1}}

\authorrunning{S. Galanakis et al.}

\institute{Imperial College London, UK \and
 Northeastern University London, UK
}




  
\maketitle

\begin{abstract}
Despite recent advances in deep generative modeling, skin lesion classification systems remain constrained by the limited availability of large, diverse, 
and well-annotated clinical datasets, resulting in class imbalance between benign and malignant lesions and consequently reduced generalization performance.
We introduce \sys{}, a rectified flow-based text-to-image generative framework that 
synthesizes clinically grounded skin lesion images from natural language descriptions of dermatological attributes.
Built upon \flux, \sys{} is fine-tuned using parameter-efficient Low-Rank Adaptation (LoRA) on a large  curated collection of publicly available clinical 
image datasets.
We construct image-text pairs using synthetic textual captions generated by Llama~3.2, following established dermatological criteria including lesion asymmetry, border irregularity, and color variation.
Extensive experiments demonstrate that \sys{} generates diverse and clinically meaningful dermatology images that improve binary classification performance by up to 6\% when augmenting small real-world datasets, and by up to 9\% when classifiers are trained on \sys{}-generated synthetic images rather than diffusion-based synthetic images.
Our ImageNet-pretrained ViT fine-tuned with only $2{,}500$ real images and $4{,}375$ \sys{}-generated samples achieves $78.04\%$ binary classification accuracy 
and an AUC of $0.859$, 
surpassing 
the next best dermatology model by $8\%$.
Code, data and pretrained models are available at \href{https://dermaflux.github.io/}{dermaflux.github.io}


\end{abstract}

\section{Introduction}

Skin cancer is among the most common cancers worldwide, posing a significant global health burden.
Globally, approximately $331{,}722$ new melanoma cases were reported in 2022~\cite{wang2025skin}, while in the United Kingdom alone, approximately $17{,}500$ new melanoma cases were reported annually, making melanoma the fifth most common cancer in the country~\cite{cancerresearchuk_melanoma_stats}.
Accurate binary classification not only ensures that malignant lesions are identified early, when treatment is most effective, but also reduces unnecessary referrals for benign lesions.

Recent advances in deep learning have demonstrated strong potential for skin lesion classification, offering scalable support for clinical decision-making.
However, deep learning models require large, well-annotated datasets that are difficult to obtain and share.
The collection of dermatological images is subject to strict ethical, legal, and privacy regulations, often requiring informed consent and institutional approval. 
Consequently, publicly accessible datasets remain limited in size and may fail to capture sufficient variability in skin tone, lesion morphology, and rare malignant subtypes.
This scarcity leads to class imbalance and reduced generalization capability in deep learning-based lesion classifiers.

Synthetic data generation is a promising solution to mitigate data scarcity. 
In this context, Stable Diffusion (SD)~\cite{Rombach_2022_CVPR} has recently been widely adopted to synthesize skin lesion images conditioned on textual descriptions~\cite{10.1007/978-3-031-53767-7_10,Farooq_2024}. 
However, two key challenges remain: (\emph{i}) reliable control over clinically meaningful lesion attributes 
during image synthesis (e.g., asymmetry, border irregularity, or color heterogeneity) and (\emph{ii}) strong semantic alignment between textual descriptions and synthesized visual features.
Kim \emph{et al.}~\cite{kim2025diffusion} incorporated visual conditioning embeddings into the diffusion model to improve conditional synthesis, while \texttt{LesionGen}~\cite{fayyad2025lesiongen} introduced 
concept-guided diffusion using structured dermatological captions to improve attribute-level prompting.
Despite these enhancements, diffusion-based lesion generators inherently rely on stochastic denoising and CLIP-based text encoders with limited context capacity.
These architectural constraints limit the efficiency and determinism of the generation process.

We propose \sys{}, a rectified flow-based framework for synthetic lesion image generation.
Unlike diffusion-based generators
that rely on stochastic 
denoising, 
rectified flows learn 
a 
deterministic transport mapping between noise and data distributions, improving sampling efficiency.
Specifically, we fine-tune \flux~\cite{blackforestlabs_flux1} via Low-Rank Adaptation (LoRA)~\cite{hu2022lora}, enabling parameter-efficient adaptation to skin lesion synthesis while preserving the pretrained model’s generative capacity.
\flux utilizes a CLIP~\cite{pmlr-v139-radford21a} and a T5-XXL~\cite{10.5555/3455716.3455856} text encoder, thus enabling richer contextual conditioning and improved text–image 
alignment.

Beyond architectural considerations, the design of open and well-annotated training datasets containing image--text pairs is equally critical for medical image synthesis.
PanDerm~\cite{yan2025multimodal} and Derm1M~\cite{yan2025derm1m} introduced large-scale datasets containing approximately one million dermatology image–text pairs.
However, a substantial proportion of PanDerm comprises in-house data that is not fully accessible to the research community, while Derm1M annotations were automatically extracted mainly from medical videos and online forums, potentially introducing noise.
In contrast, we construct an open, curated image--text dataset to train \sys{}, derived exclusively from publicly available dermatology datasets with human-verified annotations.
Textual descriptions are generated using Llama~3.2~\cite{grattafiori2024llama3herdmodels} and structured around clinically meaningful skin lesion attributes. 
Combined with the expressive dual text encoders of \flux, this structured supervision enables finer control over skin lesion characteristics such as asymmetry, border irregularity, and color variation.

Overall, our main contributions are:
\begin{enumerate}
\item 
We construct a large-scale dermatology dataset of approximately 500k image-text pairs with structured, attribute-level captions that explicitly encode skin lesion asymmetry, border irregularity, and color variation, enabling fine-grained alignment between clinically meaningful lesion descriptors and corresponding visual features.
\item 
We develop \sys{}, a rectified flow–based generative framework for synthesizing 
diverse and semantically consistent skin lesion images, offering a deterministic alternative to diffusion-based skin lesion generation.
\item 
We experimentally demonstrate that 
classifiers trained with 
\sys{}-generated
synthetic lesions 
improve binary classification performance by up to 6\%
when augmenting limited 
real-world datasets, 
achieve up to 9\% higher accuracy than diffusion-based synthetic 
augmentation under identical training settings, 
and outperform existing state-of-the-art classifiers by 8\%.
\end{enumerate}

\section{Method}

\begin{figure}[!t]
\centering
\includegraphics[width=\textwidth]{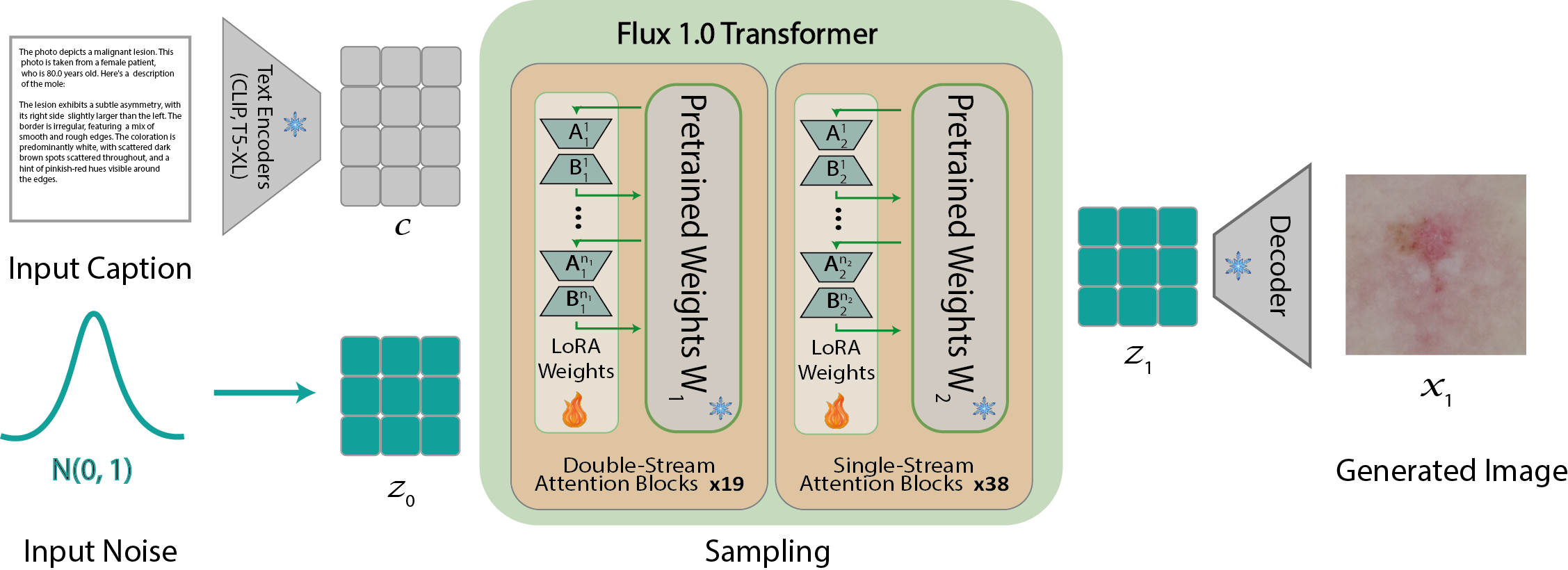}
\caption{\sys{} synthesizes a skin lesion image $x_1$ by transporting Gaussian noise $z_0$ to a clean latent representation $z_1$, conditioned on the input caption. The \flux~\cite{blackforestlabs_flux1} backbone is frozen ({\textcolor{blue}{\footnotesize \faSnowflake}}) and only the injected LoRA parameters are trained ({\textcolor{orange}{\footnotesize \faFire}}).}

\label{fig:main_figure}
\end{figure}

\sys{} is a rectified flow–based framework for text-conditioned lesion synthesis (Fig.~\ref{fig:main_figure}). 
We first present our dataset curation efforts  (\S\ref{sec:dataset}), followed by an outline of the backbone model and the adaptation strategy (\S\ref{sec:prelim}).

\subsection{Dataset Creation}
\label{sec:dataset}

\begin{table}[!t]
\centering
\caption{Summary of the aggregated dermatology datasets used for training, reporting imaging modality (clinical {\footnotesize \faCamera}, dermoscopic {\footnotesize \faSearch}, or both), 
benign and malignant counts, 
and training resolution.}
\label{tab:dataset_numbers}
{\fontsize{8pt}{10pt}\selectfont
\begin{tabular}{llrrc}
\toprule
\textbf{Dataset} & 
\textbf{Type} & 
\textbf{Benign} & 
\textbf{Malignant} & 
\textbf{Resolution} \\
\midrule
MedNode~\cite{giotis2015med_node} & 
{\footnotesize \faCamera} & 
$100$ & 
$70$ & 
$512 \times 512$ \\
HIBA~\cite{hospitalitaliano2023_skinlesions} & 
{\footnotesize \faCamera}, {\footnotesize \faSearch} & 
$802$ & 
$751$ & 
$512 \times 512$ \\
Derm12345~\cite{Yilmaz2024} & 
{\footnotesize \faSearch} & 
$11{,}120$ & 
$1{,}167$ & 
$512 \times 512$ \\
ISIC 2019~\cite{tschandl2018ham10000,codella2017isicchallenge,hernandezperez2024bcn20000} & 
{\footnotesize \faSearch} & 
$15{,}991$ & 
$9{,}340$ & 
$512 \times 512$ \\
ISIC 2020~\cite{rotemberg2021patientcentric} & 
{\footnotesize \faSearch} & 
$32{,}542$ & 
$584$ & 
$512 \times 512$ \\
Milk10k~\cite{milkstudy2025_milk10k} & 
{\footnotesize \faCamera}, {\footnotesize \faSearch} & 
$2{,}966$ & 
$7{,}514$ & 
$512 \times 512$ \\
PAD20~\cite{pacheco2020padufes20} & 
{\footnotesize \faCamera} & 
$1{,}209$ & 
$1{,}089$ & 
$512 \times 512$ \\
Kaggle 1~\cite{muhammad_hasnain_javid_2022} & 
{\footnotesize \faSearch} & 
$5{,}000$ & 
$4{,}605$ & 
$256 \times 256$ \\
Kaggle 2~\cite{kaggle_13k} & 
{\footnotesize \faSearch} & 
$6{,}289$ & 
$5{,}590$ & 
$256 \times 256$ \\
DDI~\cite{doi:10.1126/sciadv.abq6147} & 
{\footnotesize \faCamera} & 
$485$ & 
$171$ & 
$256 \times 256$ \\
ISIC 2024~\cite{isic-2024-challenge} & 
{\footnotesize \faCamera} & 
$400{,}666$ & 
$393$ & 
$128 \times 128$ \\
\bottomrule
\end{tabular}}
\end{table}

\paragraph{Dataset collection.}
We curate a large-scale dermatology image corpus by aggregating publicly available datasets for automated skin lesion analysis, including the International Skin Imaging Collaboration (ISIC) challenge datasets (2019-2024)~\cite{tschandl2018ham10000,codella2017isicchallenge,hernandezperez2024bcn20000,rotemberg2021patientcentric,milkstudy2025_milk10k}, Derm12345~\cite{Yilmaz2024} and PAD20~\cite{pacheco2020padufes20}.
Training dataset statistics are summarized in Table~\ref{tab:dataset_numbers}; 
sizes are reported before filtering.
The collection spans clinical and dermoscopic imagery, capturing substantial variability in lesion appearance, skin tone, and acquisition conditions, and thus provides a diverse foundation for synthetic skin lesion generation.
All images undergo unified pre-processing, including resolution standardization and removal of samples with minimum side length below 128 pixels. 
To accommodate varying source resolutions, the data are grouped into three training scales: $128 \times 128$, $256 \times 256$, and $512 \times 512$.
These scales contain $400{,}901$, $22{,}140$, and $85{,}303$ samples, respectively, resulting in a total dataset size of $508{,}344$ samples.

\begin{figure}[!t]
\centering
\begin{minipage}[t]{0.48\textwidth}
\begin{center}
\includegraphics[width=0.5\textwidth]{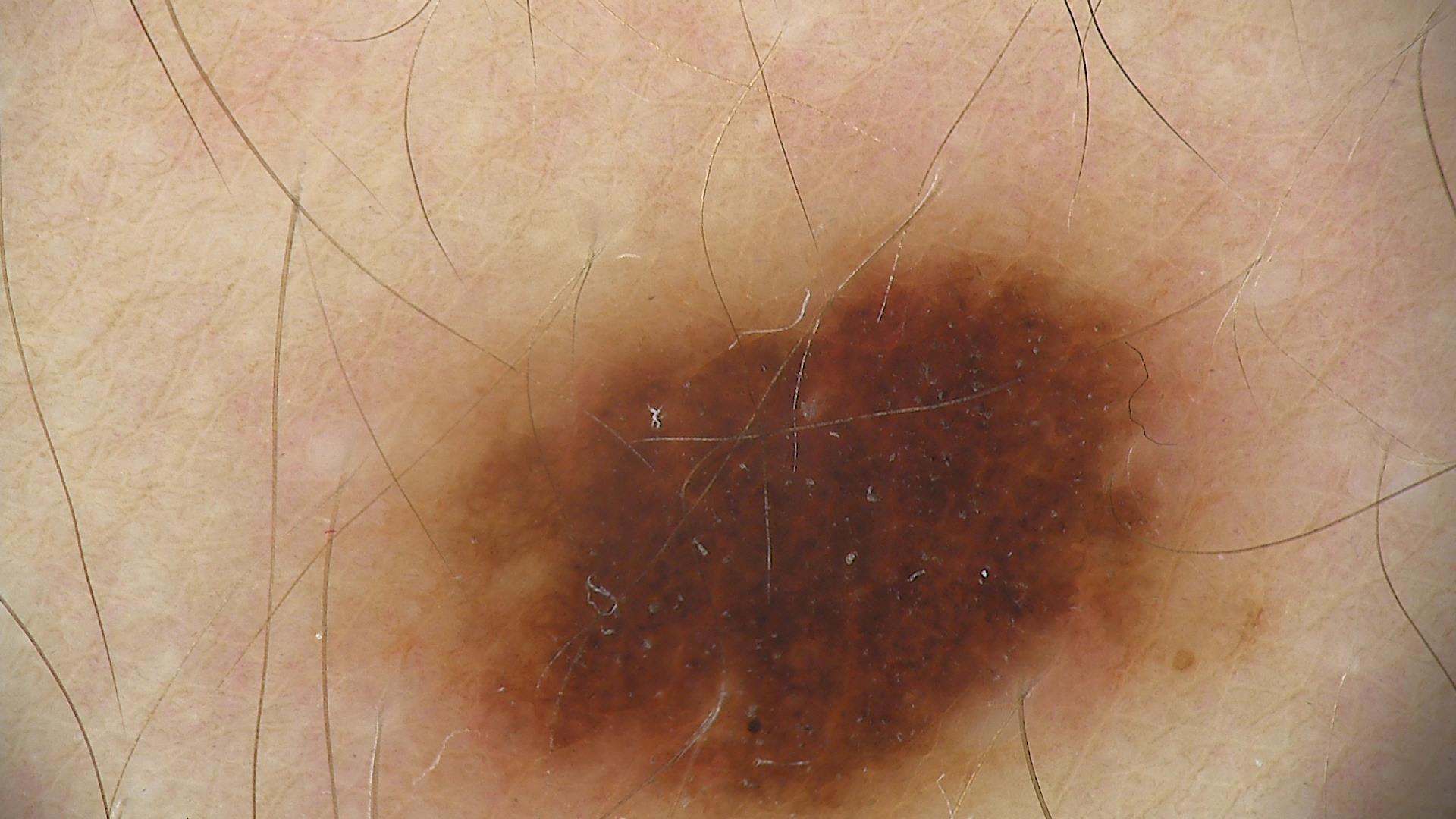}
\end{center}
{\fontsize{8pt}{10pt}\selectfont
``The mole in the image is generally 
{\sethlcolor{assymetry}
\hl{symmetrical}},
with 
{\sethlcolor{border}
\hl{a relatively smooth border. However, 
there are some areas where the border appears 
to be slightly irregular, with a few small notches
 and bumps.}}
{\sethlcolor{colour}
\hl{The color of the mole is primarily 
brown, with some areas of lighter and darker 
shades of brown. There are also some small, 
darker spots scattered throughout the mole,
 which may be pigmentation  variations or 
small freckles.}} Overall, the mole has a
 relatively uniform color and texture, with 
some minor variations.''}
\end{minipage}
\hfill
\begin{minipage}[t]{0.48\textwidth}
\begin{center}
\includegraphics[width=0.3\textwidth]{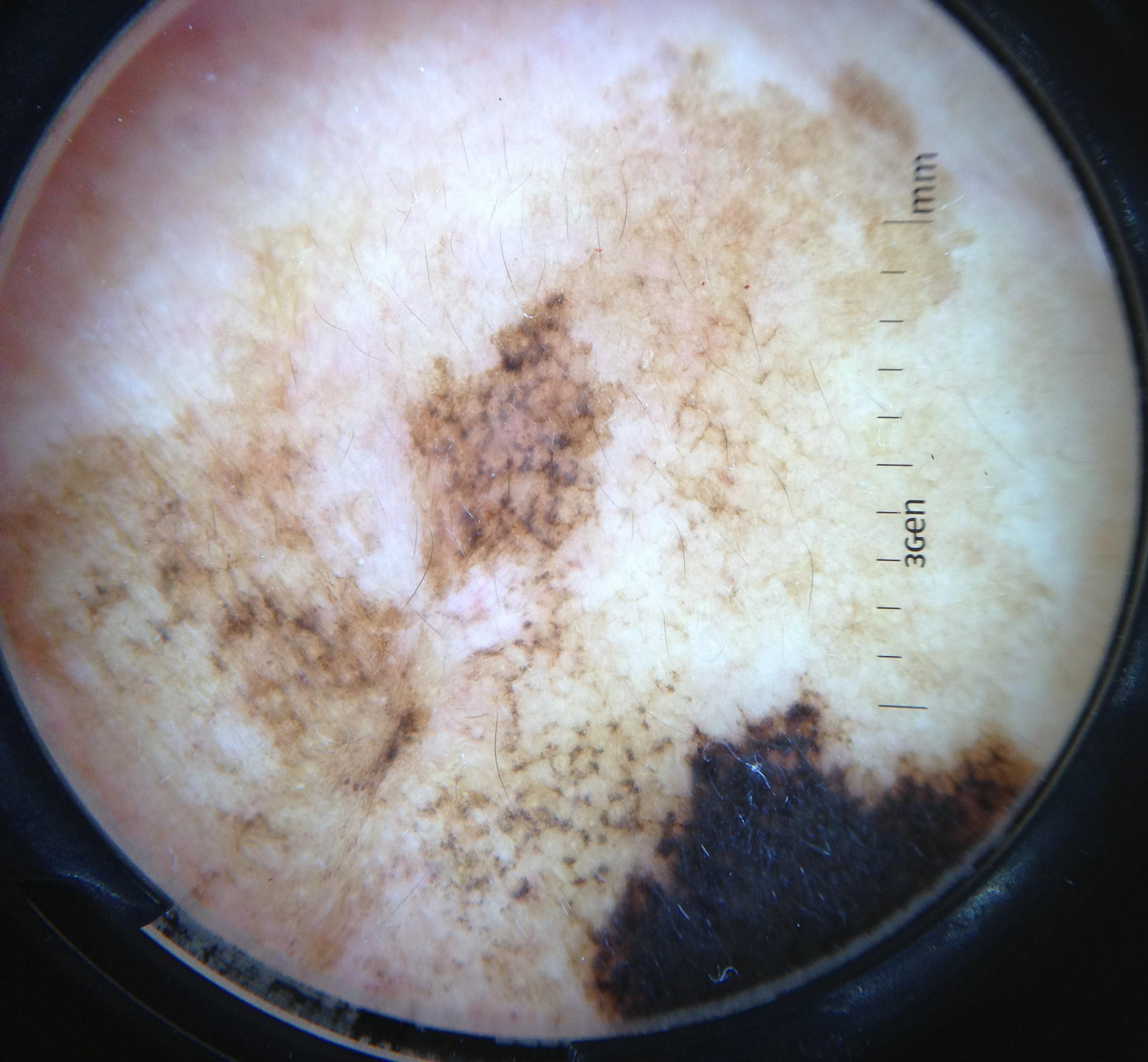}
\end{center}
{\fontsize{8pt}{10pt}\selectfont
{\sethlcolor{assymetry}
``\hl{The mole exhibits a notable degree of 
asymmetry, with one half being significantly 
larger than the other.}}
{\sethlcolor{border}
\hl{The border of the mole is
 irregular and jagged,  with a rough, uneven 
texture.}}
{\sethlcolor{colour}
\hl{The coloration of the mole is also
 irregular,  featuring a mix of dark brown and 
black patches scattered  throughout the lesion.
 Additionally, there are areas of lighter  skin 
tone visible, which may indicate a variation
 in pigmentation.}} 
Overall, the mole's
 appearance suggests a potential malignancy, 
warranting further examination  and diagnosis.''}
\end{minipage}
\caption{Given 
    a lesion image and its label, benign (\emph{left}) or malignant (\emph{right}), LLama 3.2 generates a synthetic caption using the 
    prompt: 
    ``This is an image containing a [label] lesion. 
    Give me a description of this mole regarding its
    {\sethlcolor{assymetry}
\hl{asymmetry}}, 
    {\sethlcolor{border}
\hl{border irregularity}}, 
    and {\sethlcolor{colour}
\hl{color}}.''}
\label{fig:dataset}
\end{figure}

\paragraph{Synthetic caption generation.}
Original dataset metadata provides diagnostic labels but lacks detailed morphological descriptors.
We therefore generate structured, attribute-level captions for each skin lesion image using a vision--language model (VLM).
For every image, we prompt the VLM to describe lesion characteristics following the ABC components of the ABCDE dermatological criteria---namely, asymmetry, border irregularity, and color variation. 
We evaluated three VLMs on a subset of 100 images: 
Llama~3.2~\cite{grattafiori2024llama3herdmodels}, 
ChatGPT~5~\cite{chatgpt2025}, 
and Gemini~\cite{gemini2025}. 
All models generated comparable structured descriptions, thus Llama~3.2 was selected for its scalability and local deployment capabilities.
Two clinical experts confirmed the medical plausibility of 100 captions, with representative benign and malignant examples shown in Fig.~\ref{fig:dataset}.
The final dataset contains approximately 500k structured image--text pairs 
used to train \sys{}.

\subsection{Model Architecture \& Adaptation}
\label{sec:prelim}

\sys{} adapts the pretrained \flux rectified flow backbone~\cite{blackforestlabs_flux1} to dermatological lesion synthesis through Low-Rank Adaptation (LoRA)~\cite{hu2022lora}.

\flux employs a transformer-based architecture to model rectified flows in latent space.
It comprises an encoder $\mathcal{E}$, a decoder $\mathcal{D}$, dual text encoders (CLIP~\cite{pmlr-v139-radford21a} and T5-XXL~\cite{2020t5}) and a hybrid transformer with 19 double-stream and 38 single-stream blocks.
Let $z_0 \sim p_0(z) = \mathcal{N}(\mathbf{0, I})$ denote Gaussian noise and $z_1 \sim p_1(z)$ the latent data distribution. 
Given caption embeddings $c$, rectified flows define the linear interpolation path $z_t = ( 1 - t) z_0 + t z_1, ~ t \in [0, 1],$ and learn a velocity field $\upsilon_{\theta}(z,t,c)$ that deterministically transports $z_0$ to $z_1$. 
The latent representations $z_t$ are tokenized into visual tokens and fused with text embeddings via cross-attention, enabling joint visual–semantic modeling.
The final latent $z_1$ is then decoded to the synthesized lesion image $x_1 = \mathcal{D}(z_1)$ (Fig.~\ref{fig:main_figure}).

We specialize \flux to dermatological synthesis using LoRA.
We fine-tune only the attention layers of the transformer while keeping backbone weights frozen.
In this way, we preserve the pretrained model's generative capacity while enabling 
efficient domain adaptation.
For pre-trained weights $\mathbf{W} \in \mathbb{R}^{d\times k}$, 
LoRA learns a low-rank update $\mathbf{W}' = \mathbf{W} + \frac{\alpha}{r} \mathbf{B}\mathbf{A},$ 
where 
$\mathbf{A} \in \mathbb{R}^{r\times k}$, 
$\mathbf{B} \in \mathbb{R}^{d\times r}$,
$r \ll \min(d,k)$,
and $\alpha$ scales the update.

\begin{figure}[!t]
\centering
\begin{minipage}{\linewidth}
\centering
\includegraphics[width=\textwidth]
{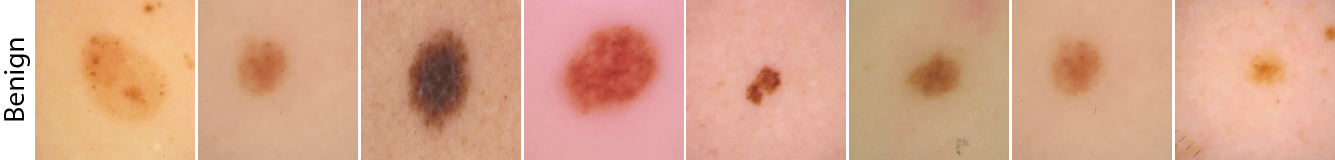}
\end{minipage}
\begin{minipage}{\linewidth}
\centering
\includegraphics[width=\textwidth]
{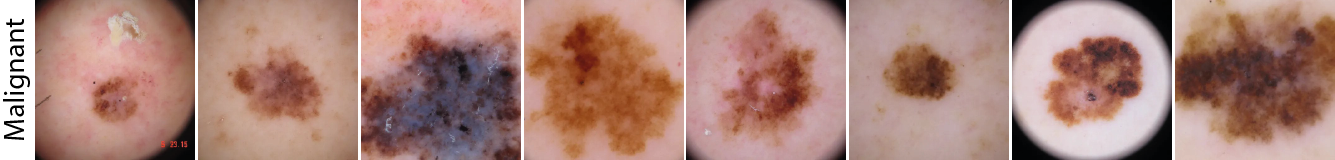}
\end{minipage}
\caption{Examples highlighting our method’s capacity to produce diverse and realistic skin lesion images in both benign (\emph{top}) and malignant (\emph{bottom}) categories.}
\label{fig:examples}
\end{figure}

We adapt \flux{}-dev (12B parameters) 
and set the LoRA hyperparameters to $r=64$ and $\alpha=64$.
Under this configuration, 
the LoRA layers amount to 
approximately 612M 
additional trainable parameters.
Training is performed for five epochs at three resolutions 
($128 \times 128$, $256 \times 256$, $512 \times 512$), 
using only the predefined training splits. 
At inference, images are generated with a flow-based ODE sampler using 20 interpolating steps. 
Fig.~\ref{fig:examples} illustrates examples of benign and malignant generated images, when we used the caption ``This is an image containing a \emph{label} lesion.'', where \emph{label} $\in$ \{benign, malignant\}.

\section{Results}

We evaluate \sys{} through a series of downstream binary classification experiments designed to assess the quality and practical utility of the generated synthetic images. 
All evaluations 
use a held-out test set of 
approximately $9{,}100$ 
benign and malignant lesions 
aggregated from the predefined 
test splits of three datasets 
in our corpus: Kaggle 1, Kaggle 2, and ISIC 2019
(Table~\ref{tab:dataset_numbers}).
We train two representative classifiers, 
ResNeXt~\cite{xie2017aggregatedresidualtransformationsdeep} and Vision Transformer (ViT)~\cite{dosovitskiy2020vit}, to ensure model architecture diversity.
Specifically, 
we use \text{ResNeXt-50 (32$\times$4d)},
initialized with ImageNet-1K pretrained weights,
and
\text{ViT-Base (Patch16-224)},
initialized with ImageNet-21K pretrained weights.
Both classifiers are trained for 40 epochs with batch size 32.
Base learning rates are $1\times10^{-4}$ for ResNeXt and $2\times10^{-5}$ for ViT,
with 
a cosine annealing schedule. 

First, we compare \sys{} against the diffusion-based Derm-T2IM~\cite{Farooq_2024} 
under controlled real-to-synthetic training ratios to assess relative synthetic image quality (\S\ref{sec:dermT2IM}).
Second, we perform a sensitivity analysis by varying the proportion of synthetic data to quantify its contribution to model generalization (\S\ref{sec:sensitivity}). 
Finally, we benchmark our best-performing configuration against state-of-the-art 
dermatology classifiers (\S\ref{sec:comparison}).

\begin{table}[!t]
    \centering
    \begin{minipage}[t]{0.58\textwidth}
    {\fontsize{8pt}{10pt}\selectfont
    \caption{Classification accuracy of ResNeXt \\ and ViT under our two training scenarios.}
    \label{tab:dermt2i_acc}
    \centering
   \begin{tabular}{lccc}
\toprule
& Scenario & Derm-T2IM & \sys \\
\midrule
\multirow{2}{*}{{ResNeXt}}
 & (\emph{i}) & 0.6278 & \textbf{0.6398} \\
 & (\emph{ii}) & 0.6412 & \textbf{0.6525} \\
\midrule
\multirow{2}{*}{{ViT}}
 & (\emph{i}) & 0.6795 & \textbf{0.7698} \\
 & (\emph{ii}) & 0.7091 & \textbf{0.7438} \\
\bottomrule
\end{tabular}
    }
    \end{minipage}
    \hfill
    \begin{minipage}[t]{0.4\textwidth}
    {\fontsize{8pt}{10pt}\selectfont
    \caption{Comparison with state-of-the-art dermatology models.}
    \label{tab:acc_other_sota}
    \begin{tabular}{lc}
    \toprule
    \textbf{Method}  & \textbf{Accuracy} \\
    \midrule
    BiomedCLIP~\cite{zhang2024biomedclip} & 0.5474 \\
    DermLIP-ViT~\cite{yan2025derm1m} & 0.6603 \\
    MAKE~\cite{yan2025makemultiaspectknowledgeenhancedvisionlanguage} & 0.7012 \\
    DermLIP-PanDerm~\cite{yan2025derm1m} & 0.7047\\ 
    \midrule
    \sys{}-ViT(Ours)  & \textbf{0.7804} \\
    \bottomrule
    \end{tabular}
    }
    \end{minipage}
\end{table}

\subsection{Downstream Evaluation of Synthetic Image Quality}
\label{sec:dermT2IM}

We assess the quality of synthetic images generated by \sys{} through a controlled downstream comparison with Derm-T2IM~\cite{Farooq_2024}, 
a synthetic lesion generator built upon SD~\cite{Rombach_2022_CVPR}.
Derm-T2IM is designed to enhance dermatology classification in data-scarce settings and $6{,}000$ synthetic lesions are publicly available.
We consider two training scenarios:
\begin{inparaenum}[\it (i)]
\item classifiers trained 
exclusively on synthetic data:
$6{,}000$ images from Derm-T2IM
and
$6{,}000$ from \sys{};
\item 
classifiers trained 
on a mixture of $2{,}500$ 
real images and 
$5{,}000$ synthetic images 
from each method, with Derm-T2IM samples 
randomly drawn from 
the authors' released dataset.
\end{inparaenum}
In both scenarios, datasets are balanced between benign and malignant skin lesions;
and ResNeXt and ViT are trained under identical training settings.

Across both architectures and scenarios, classifiers trained on \sys{}-generated samples consistently outperform those trained on diffusion-generated samples.
Table~\ref{tab:dermt2i_acc} shows that ResNeXt achieves gains of $+1.20\%$ and $+1.13\%$ in scenarios (\emph{i}) and (\emph{ii}), respectively.
The improvement is more pronounced for ViT, gaining $+9.03\%$ and $+3.47\%$ 
under the same settings. 
These results indicate superior downstream utility of the proposed synthetic image generator.
 
\begin{figure}[!t]
\centering
\begin{minipage}[c]{0.55\linewidth}
    \centering
    \includegraphics[width=\textwidth]{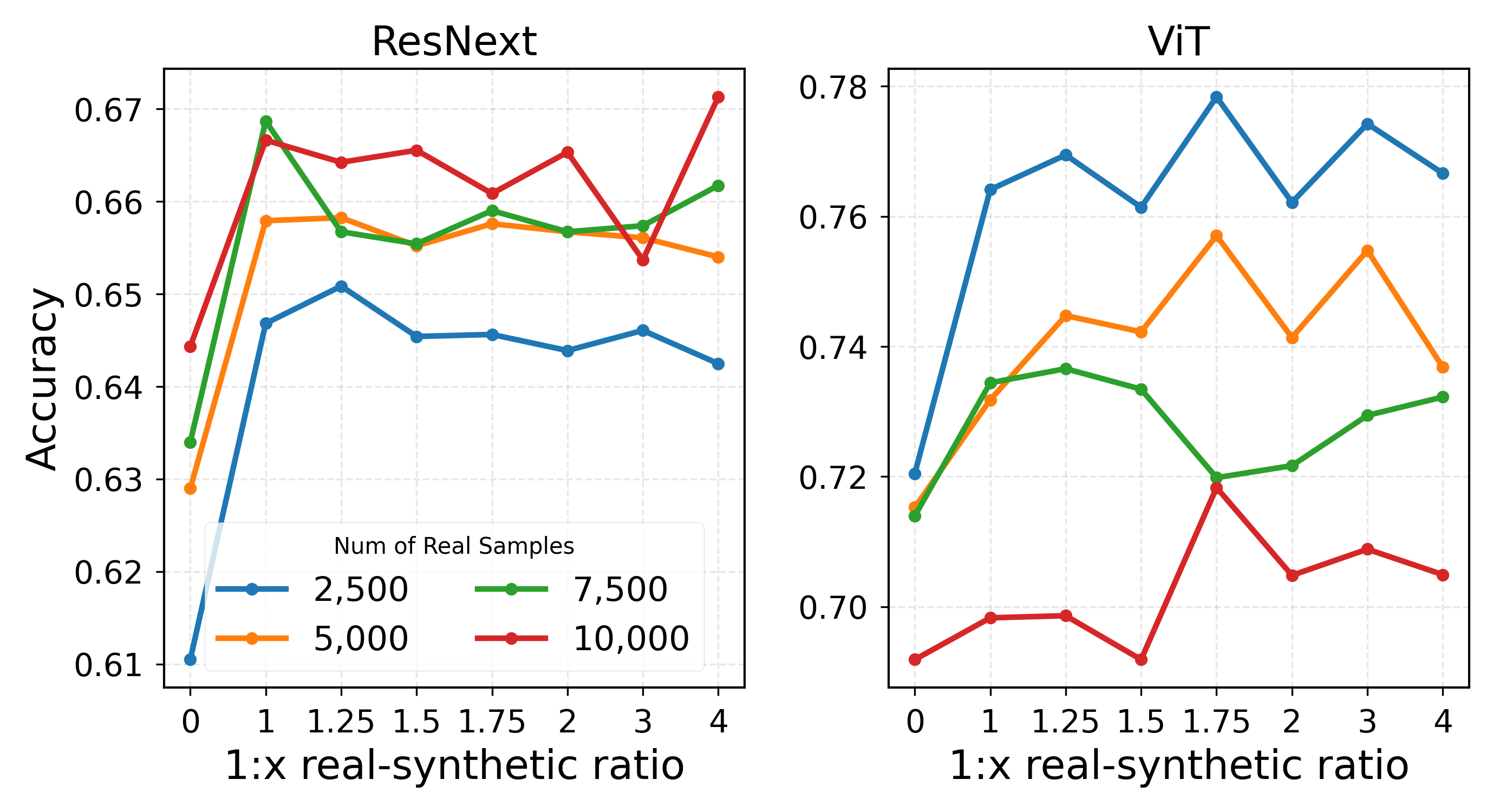}
    \caption{Test accuracy scores for ResNeXt and ViT across varying real-to-synthetic training data ratios.
    Results are averaged 
over five independent runs (different seeds). }
    \label{fig:class_score}
\end{minipage}
\begin{minipage}[c]{0.4\textwidth}
    \centering 
    \includegraphics[width=0.86\textwidth]{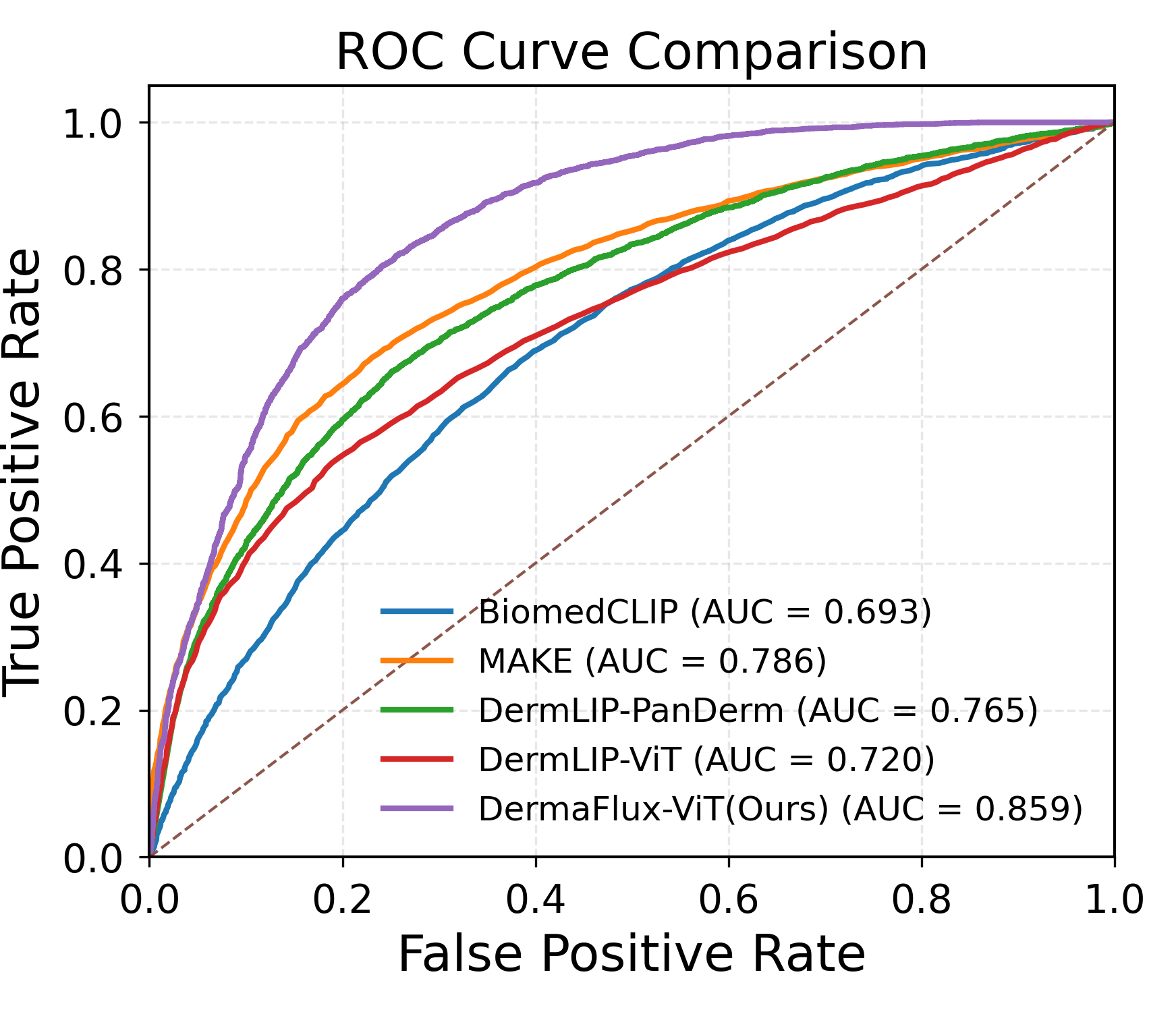}
    \caption{\sys{}-ViT separates malignant and benign test samples more 
    reliably than competing models.
    }
    \label{fig:roc_curves}
\end{minipage}
\end{figure}

\subsection{Impact of Synthetic Data Augmentation}
\label{sec:sensitivity}

We evaluate synthetic data augmentation by progressively enriching balanced real-image training sets with \sys{}-generated samples. 
For each experiment, a fixed number of real images 
($2{,}500$, $5{,}000$, $7{,}000$, $10{,}000$) 
is randomly sampled from our dataset 
with 
equal numbers of benign and malignant cases. 
Synthetic images are then incrementally added to construct varying real-to-synthetic ratios.
Ratios are expressed as $1$:$x$, meaning that for every $n$ real images, $n\cdot x$ synthetic samples are added to the training set.
We train ResNeXt and ViT 
and evaluate performance using Accuracy and ROC-AUC scores.

Synthetic augmentation improves generalization by enriching data diversity, particularly when real data are limited.
Fig.~\ref{fig:class_score} shows that, 
across both architectures, incorporating synthetic data consistently improves performance over real-only training. 
The largest gains occur when moving to balanced real–synthetic configurations 
(e.g., 1:1).
For example, with $5{,}000$ real samples, ResNeXt improves from 63.2\% to 64.68\% 
accuracy when synthetic samples are added, while ViT increases from 71.3\% to 76.1\%.
Improvements are more pronounced 
in lower-data regimes 
($2{,}500$ and $5{,}000$ real samples) 
and remain positive but smaller for larger training sets.
We additionally evaluate
purely synthetic training 
(0:1 ratio; $2{,}500$ and $5{,}000$ samples).
Even without real images, 
\sys{} improves accuracy 
by 0.7-2.3\% for ResNeXt 
and 1.84-4.66\% for ViT 
compared to the 1:0 ratio,
suggesting that generated 
samples capture 
class-discriminative 
lesion characteristics 
beyond simple augmentation effects.

Consistent improvements 
in ROC-AUC mirror 
the accuracy trends:
e.g., gains of $+3.43\%$
and $+3.00\%$ for ResNeXt 
and $+1.97\%$ and $+4.68\%$ 
for ViT 
when moving 
from a 1:0 to a 1:1 ratio 
with $2{,}500$ and $5{,}000$ 
real images, respectively.
Since ROC-AUC measures 
the probability that a malignant lesion receives a higher confidence score than a benign one, 
these improvements
indicate
that training with 
\sys{}-generated 
images
enhances
class separability. 

\subsection{Comparison with State-of-the-Art Classifiers}
\label{sec:comparison}

We evaluate the effectiveness of the proposed synthetic data generation framework by comparing our trained ViT model against state-of-the-art dermatology classifiers: 
BiomedCLIP~\cite{zhang2024biomedclip}, 
MAKE~\cite{yan2025makemultiaspectknowledgeenhancedvisionlanguage}, 
and the DermLIP variants introduced in Derm1M~\cite{yan2025derm1m}. 
Table~\ref{tab:acc_other_sota} reports the binary classification accuracy for each method on the held-out test set.

Across all evaluated training 
configurations, 
\sys{}-ViT consistently 
outperforms prior state-of-the-art 
dermatology classifiers.
Using $2{,}500$ real images 
and varying the real-to-synthetic ratio, 
accuracy ranges from $76\%$ to $78.04\%$.
The best performance is 
achieved with $4{,}375$ synthetic samples, 
reaching $78.04\%$ accuracy---an 
improvement of 
$8$ percentage points 
over MAKE and DermLIP.
These findings indicate 
that clinically structured 
synthetic data can 
effectively compensate 
for limited real-world 
supervision, enabling 
competitive performance 
with substantially reduced 
training data.

Figure~\ref{fig:roc_curves} 
illustrates the 
discriminative capability 
of \sys{}-ViT 
via ROC analysis.
Our best-performing ViT
achieves the highest AUC ($0.859$)
among all compared models.
Notably,
its ROC curve approaches 
a true positive rate (TPR) 
close to $1.0$ 
at a 
false positive rate (FPR)
of $0.75$, 
while prior models 
do not reach similar sensitivity levels.
This operating characteristic 
enables 
\sys{}-ViT to achieve 
very high sensitivity 
while maintaining 
higher specificity.
From a clinical perspective, 
this 
reduces both 
the risk 
of missed malignant lesions 
and the number of
unnecessary 
referrals and 
follow-up examinations.

\section{Conclusion}

We introduced \sys{}, a rectified flow--based text-to-image framework for dermatological lesion synthesis.
By fine-tuning \flux with LoRA and conditioning on structured Llama~3.2 captions, \sys{} produces clinically coherent lesion images with improved attribute control and text--image alignment.
In downstream evaluation, \sys{} improves binary classification accuracy by up to $6\%$ when augmenting limited real datasets and by up to $9\%$ over diffusion-based synthetic images.
With only $2{,}500$ real images and $4{,}375$ synthetic samples, \sys{}-ViT achieves $78.04\%$ accuracy and an AUC of $0.859$, 
exceeding the next best dermatology model by $8\%$.
Overall, 
clinically structured synthetic data 
reduce reliance on large-scale 
dataset curation 
while improving generalization 
and class separability 
in data-scarce settings.

\subsection*{Acknowledgments}
This work was supported by the EPSRC Turing AI Fellowship (Grant Ref: EP/Z534699/1): Generative Machine
Learning Models for Data of Arbitrary Underlying Geometry (MAGAL).
%
%
%
\bibliographystyle{splncs04}
\bibliography{bibliography}

\end{document}